\documentclass[final]{cvpr}

\usepackage{times}
\usepackage{epsfig}
\usepackage{graphicx}
\usepackage{amsmath}
\usepackage{amssymb,booktabs,tabulary,multirow}
\usepackage[caption=false]{subfig}

\newcommand{\tabincell}[2]{\begin{tabular}{@{}#1@{}}#2\end{tabular}}
\newcolumntype{x}[1]{>{\centering\arraybackslash}p{#1pt}}
\newlength\savewidth\newcommand\shline{\noalign{\global\savewidth\arrayrulewidth
  \global\arrayrulewidth 1pt}\hline\noalign{\global\arrayrulewidth\savewidth}}
\newcommand{\tablestyle}[2]{\setlength{\tabcolsep}{#1}\renewcommand{\arraystretch}{#2}\centering\footnotesize}
\makeatletter\renewcommand\paragraph{\@startsection{paragraph}{4}{\z@}
  {.5em \@plus1ex \@minus.2ex}{-.5em}{\normalfont\normalsize\bfseries}}\makeatother

\usepackage[pagebackref=true,breaklinks=true,colorlinks,bookmarks=false]{hyperref}

\begin{document}

\title{You Only Look One-level Feature}

\author{
Qiang Chen$^{1,2}$\thanks{This work is done during Qiang Chen's internship at MEGVII Technology.} , Yingming Wang$^4$, Tong Yang$^4$, Xiangyu Zhang$^4$, Jian Cheng$^{1,2,3}$\thanks{Corresponding author.} , Jian Sun$^4$\\
$^1$NLPR, Institute of Automation, Chinese Academy of Sciences\\
$^2$School of Artificial Intelligence, University of Chinese Academy of Sciences\\
$^3$CAS Center for Excellence in Brain Science and Intelligence Technology\\
$^4$MEGVII Technology\\
{\tt\small \{qiang.chen, jcheng\}@nlpr.ia.ac.cn, \{wangyingming, yangtong, zhangxiangyu, sunjian\}@megvii.com}
}

\maketitle

\begin{abstract}
This paper revisits feature pyramids networks (FPN) for one-stage detectors and points out that the success of FPN is due to its divide-and-conquer solution to the optimization problem in object detection rather than multi-scale feature fusion. From the perspective of optimization, we introduce an alternative way to address the problem instead of adopting the complex feature pyramids - {\em utilizing only one-level feature for detection}. Based on the simple and efficient solution, we present You Only Look One-level Feature (YOLOF). In our method, two key components, Dilated Encoder and Uniform Matching, are proposed and bring considerable improvements. Extensive experiments on the COCO benchmark prove the effectiveness of the proposed model. Our YOLOF achieves comparable results with its feature pyramids counterpart RetinaNet while being $2.5\times$ faster. Without transformer layers, YOLOF can match the performance of DETR in a single-level feature manner with $7\times$ less training epochs. With an image size of $608\times608$, YOLOF achieves 44.3 mAP running at 60 fps on 2080Ti, which is $13\%$ faster than YOLOv4. Code is available at  \url{https://github.com/megvii-model/YOLOF}.
\end{abstract}

\section{Introduction} \label{sec1}
In state-of-the-art two-stage detectors~\cite{lin2017feature,he2017mask,cai2018cascade} and one-stage detectors~\cite{lin2017focal,tian2019fcos}, feature pyramids become an essential component. The most popular way to build feature pyramids is the feature pyramid networks (FPN)~\cite{lin2017feature}, which mainly brings two benefits: (1) {\em multi-scale feature fusion}: fusing multiple low-resolution and high-resolution feature inputs to obtain better representations; (2) {\em divide-and-conquer}: detecting objects on different levels regarding objects' scales. A common belief for FPN is that its success relies on the fusion of multiple level features, inducing a line of studies of designing complex fusion methods manually~\cite{liu2018path,kong2018deep,pang2019libra}, or via Neural Architecture Search (NAS) algorithms~\cite{ghiasi2019fpn,tan2020efficientdet}. However, the belief ignores the function of the divide-and-conquer in FPN. It leads to fewer studies on how these two benefits contribute to FPN's success and may hinder new advances. 
\begin{figure}
\centering
\includegraphics[width=.45\textwidth]{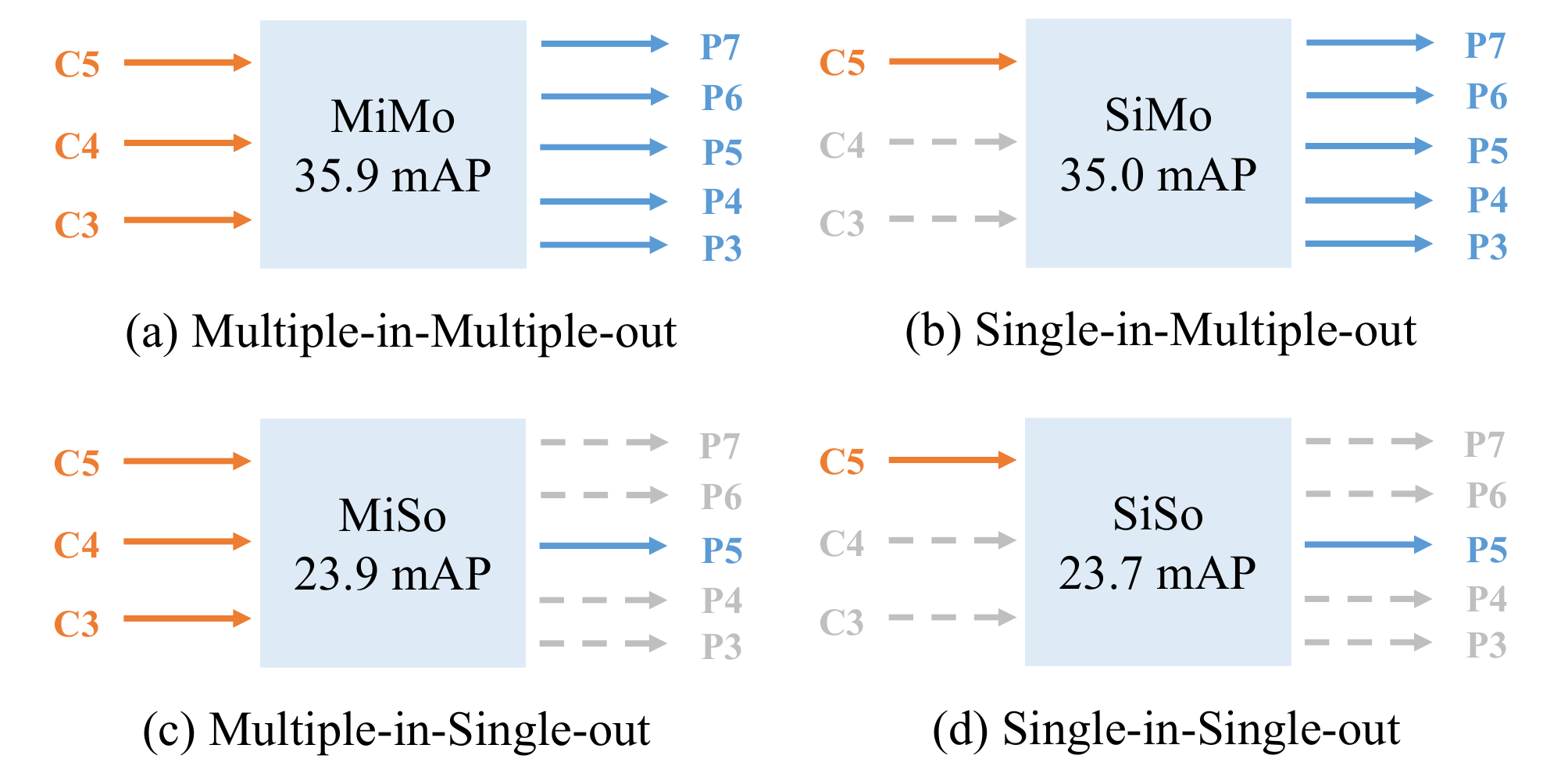}
\caption{Comparison of box AP among the Multiple-in-Multiple-out (MiMo), Single-in-Multiple-out (SiMo), Multiple-in-Single-out (MiSo), and Single-in-Single-out (SiSo) encoders on COCO validation set. Here, we adopt the original RetinaNet~\cite{lin2017focal} as our baseline model, where C3, C4, and C5 denote output features of the backbone with a downsample rate of $\{8, 16, 32\}$ and P3 to P7 represent the feature levels used for final detection. All results reported in the figure use the same backbone, ResNet-50~\cite{he2016deep}. The structure of MiMo is same as the FPN in RetinaNet~\cite{lin2017focal}. A detailed illustration of the structure for all encoders can be found in the Figure~\ref{fig9}.}
\label{fig1}\vspace{-3mm}
\end{figure}

This paper studies the influence of FPN's two benefits in one-stage detectors. We design experiments by decoupling the {\em multi-scale feature fusion} and the {\em divide-and-conquer} functionalities with RetinaNet~\cite{lin2017focal}. In detail, we consider FPN as a {\em Multiple-in-Multiple-out} (MiMo) encoder, which encodes multi-scale features from the backbone and provides feature representations for the decoder (the detection heads). We conduct controlled comparisons among {\em Multiple-in-Multiple-out} (MiMo), {\em Single-in-Multiple-out} (SiMo), {\em Multiple-in-Single-out} (MiSo), and {\em Single-in-Single-out} (SiSo) encoders in Figure~\ref{fig1}. Surprisingly, the SiMo encoder, which only has one input feature C5 and does not perform feature fusion, can achieve comparable performance with the MiMo encoder (i.e., FPN). The performance gap is less than $1$ mAP. In contrast, the performance drops dramatically ($\geq12$ mAP) in MiSo and SiSo encoders.  These phenomenons suggest two facts: (1) the C5 feature carries sufficient context for detecting objects on various scales, which enables the SiMo encoder to achieve comparable results; (2) the {\em multi-scale feature fusion} benefit is far away less critical than the {\em divide-and-conquer} benefit, thus {\em multi-scale feature fusion might not be the most significant benefit of FPN}, which is also demonstrated by ExFuse~\cite{zhang2018exfuse} in semantic segmentation. Thinking one step deeper, {\em divide-and-conquer} is related to the optimization problem in object detection. It divides the complex detection problem into several sub-problems by object scales, facilitating the optimization process. 

The above analysis suggests that the essential factor for the success of FPN is its solution to the optimization problem in object detection. The {\em divide-and-conquer} solution is a good way. But it brings memory burdens, slows down the detectors, and make detectors' structure complex in one-stage detectors like RetinaNet~\cite{lin2017focal}. Given that the C5 feature carries sufficient context for detection, we show a simple way to address the optimization problem.

We propose {\em You Only Look One-level Feature (YOLOF)}, which only uses one single C5 feature (with a downsample rate of 32) for detection. To bridge the performance gap between the SiSo encoder and the MiMo encoder, we first design the structure of the encoder properly to extract the multi-scale contexts for objects on various scales, compensating for the lack of multiple-level features; then, we apply a uniform matching mechanism to solve the imbalance problem of positive anchors raised by the sparse anchors in the single feature. 

Without bells and whistles, YOLOF achieves comparable results with its feature pyramids counterpart RetinaNet~\cite{lin2017focal} but $2.5\times$ faster. In a single feature manner, YOLOF matches the performance of the recent proposed DETR~\cite{carion2020end} while converging much faster ($7\times$). With an image size of $608\times608$ and other techniques~\cite{bochkovskiy2020yolov4,zhang2020swa}, YOLOF achieve $44.3$ mAP running at $60$ fps on 2080Ti, which is $13\%$ faster than YOLOv4~\cite{bochkovskiy2020yolov4}. In a nutshell, the contributions of this paper are:
\begin{itemize}
    \item We show that the most significant benefits of FPN is its {\em divide-and-conquer} solution to the optimization problem in dense object detection rather than the {\em multi-scale feature fusion}. 
    \item We present YOLOF, which is a simple and efficient baseline without using FPN. In YOLOF, we propose two key components, {\em Dilated Encoder} and {\em Uniform Matching}, bridging the performance gap between the SiSo encoder and the MiMo encoder.
    \item  Extensive experiments on COCO benchmark indicates the importance of each component. Moreover, we conduct comparisons with RetinaNet~\cite{lin2017focal},  DETR~\cite{carion2020end} and YOLOv4~\cite{bochkovskiy2020yolov4}. We can achieve comparable results with a faster speed on GPUs.
\end{itemize}
\begin{figure*}
\centering
\includegraphics[width=.9\textwidth]{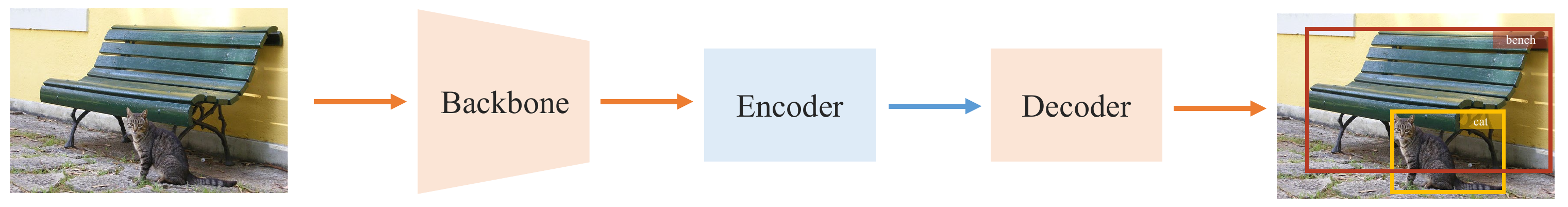}
\caption{An illustration of the detection pipeline. In this paper, we format the detection pipeline into three parts: (1) the backbone; (2) the encoder, which receives inputs from the backbone and distributes representations for detection; (3) the decoder, which performs classification and regression tasks and generate final prediction boxes. The color for the encoder is corresponding to the one in Figure~\ref{fig1}.}
\label{fig2}\vspace{-1mm}
\end{figure*}

\section{Related Works} \label{sec2}
\paragraph{Multiple-level feature detectors.} It is a conventional technique to employ multiple features for object detection. Typical approaches to construct multiple features can be categorized into image pyramid methods and feature pyramid methods. Image pyramids based detector such as DPM~\cite{felzenszwalb2009object} dominates the detection in the pre-deep learning era. In CNN-based detectors, the image pyramids method also wins some researchers'~\cite{singh2018analysis,singh2018sniper} praise as it can achieve higher performance out of the box. However, the image pyramids method is not the only way to obtain multiple features; it is more efficient and natural to exploit feature pyramids' power in CNN models. SSD~\cite{liu2016ssd} first utilizes multiple-scale features and performs object detection on each scale for different scales objects. FPN~\cite{lin2017feature} follows SSD~\cite{liu2016ssd} and UNet~\cite{ronneberger2015u} and constructs semantic-riched feature pyramids by combining shallow features and deep features. After that, several works~\cite{kong2018deep,liu2018path,ghiasi2019fpn,tan2020efficientdet} follow FPN and focus on how to obtain better representations. FPN becomes an essential component and dominates modern detectors. It is also applied to popular one-stage detectors, such as RetinaNet~\cite{lin2017focal}, FCOS~\cite{tian2019fcos}, and their variants~\cite{zhang2020bridging}. Another line of method to get feature pyramids is to use multi-branch and dilation convolution~\cite{li2019scale}. Different from the above works, our method is a single-level feature detector.

\paragraph{Single-level feature detectors.} In early times, the R-CNN series~\cite{girshick2014rich,girshick2015fast,ren2015faster} and R-FCN~\cite{dai2016r} only extract RoI features on a single feature, while their performances lag behind their multiple feature counterparts~\cite{lin2017feature}. Also, in one-stage detectors, YOLO~\cite{redmon2016you} and YOLOv2~\cite{redmon2017yolo9000} only use the last output feature of the backbone. They can be super fast but have to bear a performance decline in detection. CornerNet~\cite{law2018cornernet} and CenterNet~\cite{zhou2019objects,duan2019centernet} follow this fashion and achieve competitive results while using a single feature with a downsample rate of 4 to detect all the objects. Using a high-resolution feature map for detection brings enormous memory cost and is not friendly to deployment. Recently, DETR~\cite{carion2020end} introduces the transformer~\cite{vaswani2017attention} to detection and shows that it could achieve state-of-the-art results only use a single C5 feature. Due to the totally anchor-free mechanism and transformer learning phase, DETR needs a long training schedule for its convergence. The long training schedule characteristic is cumbersome for further improvements. Unlike these papers, we investigate the working mechanism of multiple-level detection. From the perspective of optimization, we provide an alternative solution to the widely used FPN. Moreover, YOLOF converges faster and achieves promising performance; thus, YOLOF can serve as a simple baseline for fast and accurate detectors. 
\section{Cost Analysis of MiMo Encoders}
As mentioned in Section~\ref{sec1}, the success of FPN in dense object detection is due to its solution to the optimization problem. However, the multi-level feature paradigm is inevitable to make detectors complex, brings memory burdens, and slows down the detector. In this section, we provide a quantitative study on the cost of MiMo encoders.

\begin{figure}
\centering
\includegraphics[width=.44\textwidth]{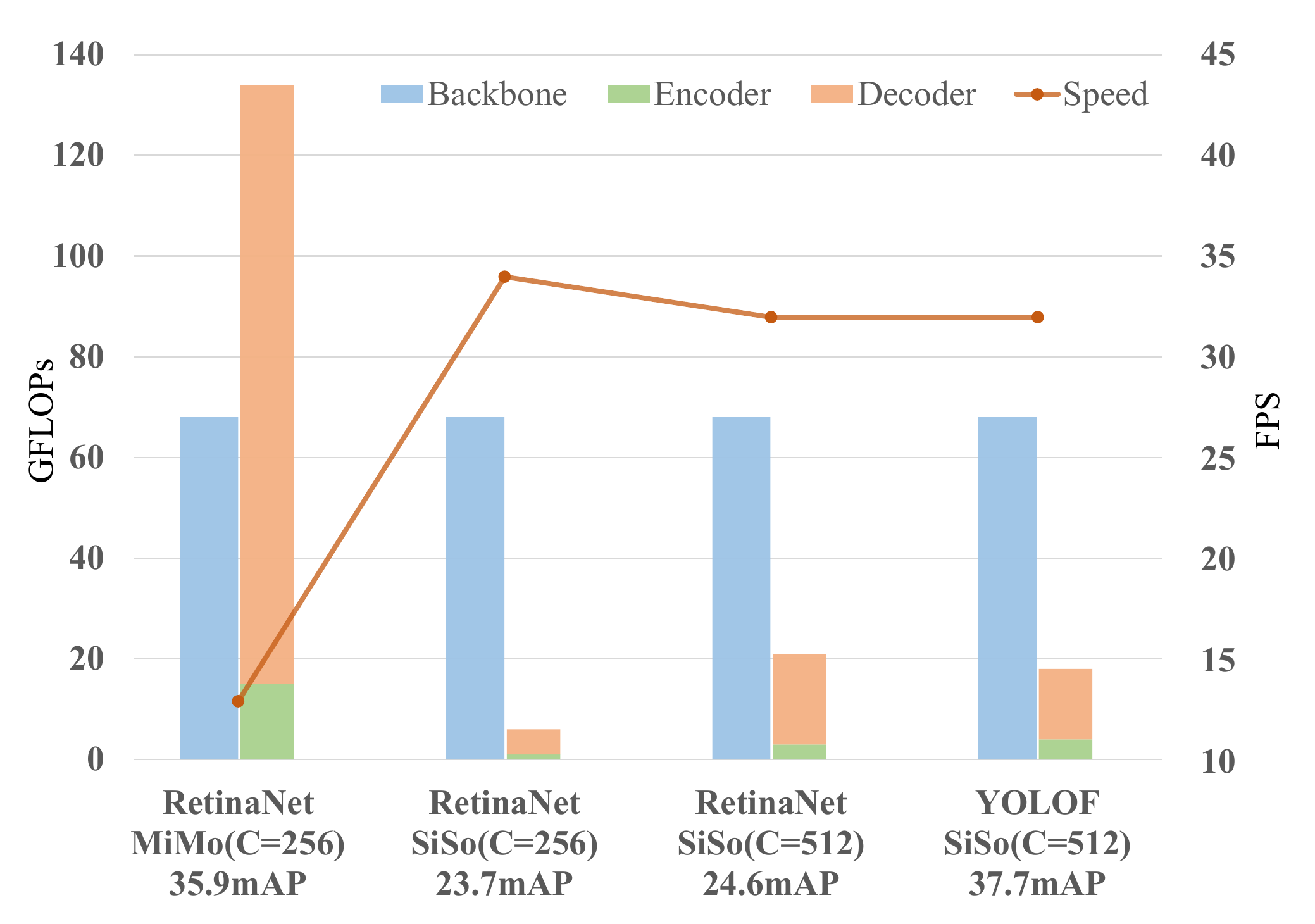}
\caption{FLOPs, accuracy, and speed comparison between the models that adopt MiMo and SiSo encoders on COCO. As the FLOPs of the decoder is affected by the encoder's outputs, we stack the FLOPs of the encoder and the decoder in the figure to better understanding the effects of encoders on the FLOPs. All models use the same backbone, ResNet-50. All FLOPs are measured with a shorter edge size 800 over the first 100 images of COCO val2017. The FPS is calculated with batch size 1 on 2080Ti from the total inference pure compute time reported in the Detectron2~\cite{wu2019detectron2}. In the figure, {\em C} represents the number of channels used in the model's encoder and decoder.}
\label{fig3}\vspace{-3mm}
\end{figure}

We design experiments based on RetinaNet~\cite{lin2017focal} with ResNet-50~\cite{he2016deep}. In detail, we format the pipeline for the detection task as a combination of three key parts: the backbone, the encoder, and the decoder (Figure~\ref{fig2}). In this view, we show the FLOPs of each component in Figure~\ref{fig3}. Compared with SiSo encoders, the MiMo encoder brings enormous memory burdens to the encoder and the decoder(134G vs. 6G) (Figure~\ref{fig3}). Moreover, the detector with MiMo encoder runs much slower than the ones with SiSo encoders (13 FPS vs. 34 FPS) (Figure~\ref{fig3}). The slow speed is caused by detecting objects on high-resolution feature maps in the detector with MiMo encoder, such as the C3 feature (with a downsample rate of 8). Given the above drawbacks of the MiMo encoder, we aim to find an alternative way to solve the optimization problem while keeping the detector simple, accurate, and fast simultaneously.

\section{Method}
Motivated by the above purpose and the finding that the C5 feature contains enough context for detecting numerous objects, we try to replace the complex MiMo encoder with the simple SiSo encoder in this section. But this replacement is \textbf{nontrivial} as the performance drops extensively when applying SiSo encoders according to the results in Figure~\ref{fig3}. Given the situation, we carefully analyze the obstacles preventing SiSo encoders from getting a comparable performance with MiMo encoders. We find that two problems brought by SiSo encoders are responsible for the performance drop. The first problem is that {\em the range of scales matching to the C5 feature's receptive field is limited}, which impedes the detection performance for objects across various scales. The second one is {\em the imbalance problem on positive anchors} raised by sparse anchors in the single-level feature. Next, we discuss these two problems in detail and provide our solutions.

\subsection{Limited Scale Range} \label{sec4.1}
\begin{figure}
\centering
\includegraphics[width=.475\textwidth]{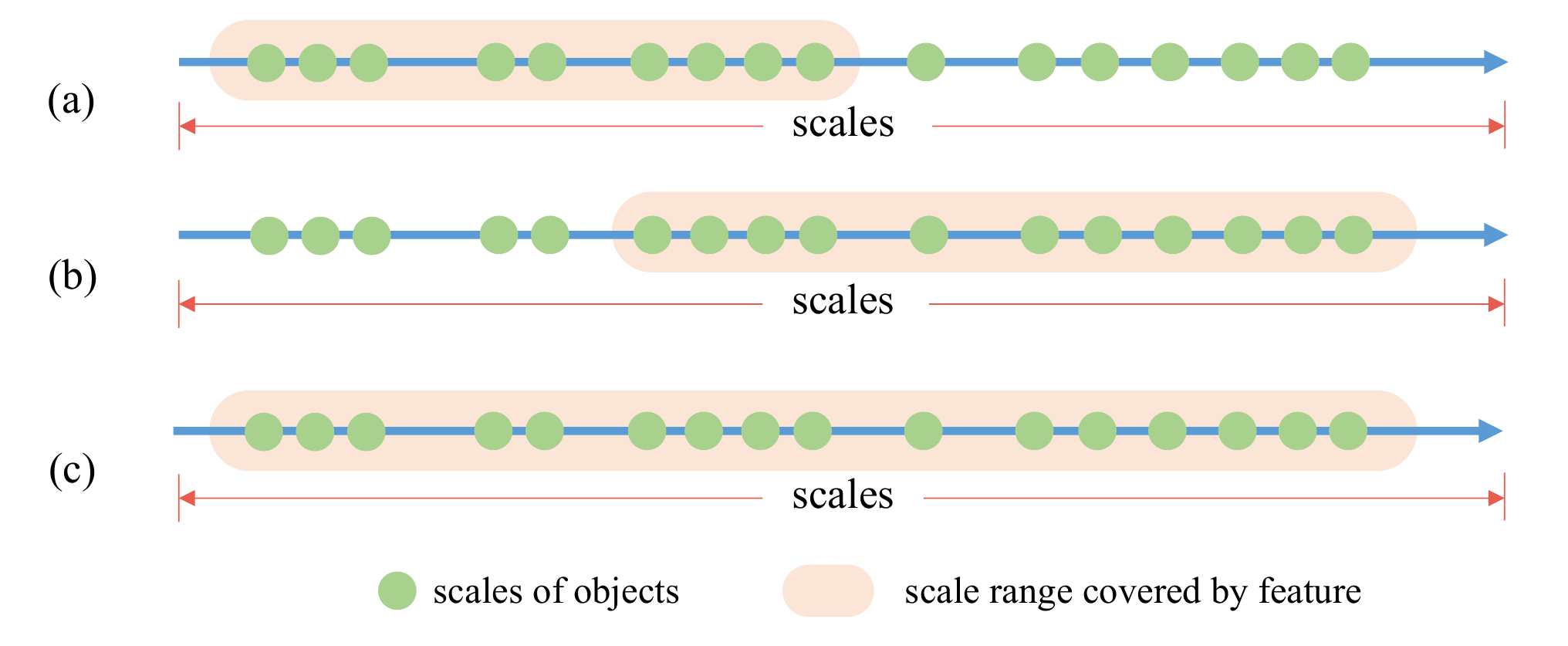}
\caption{A toy example to illustrate the relation between the object scales and the scale range covered by the single feature. The axis in this figure denotes the scales. (a) indicates that the feature's receptive field can only cover a limited scale range; (b) shows that the enlarged scale ranges enable the feature to cover large objects while miss covering small ones; (c) represents that all scales can be covered the feature with multiple receptive fields.}
\label{fig4}\vspace{-2mm}
\end{figure}

Recognizing objects at vastly different scales is a fundamental challenge in object detection. One feasible solution to this challenge is to leverage multiple-level features. In detectors with MiMo or SiMo encoders,  they construct multiple-level features with different receptive fields (P3-P7) and detect objects on the level with receptive field matching to their scales. However, the single-level feature setting changes the game. There is only one output feature in SiSo encoders, whose receptive field is a constant. As shown in Figure~\ref{fig4}(a), the C5 feature's receptive field can only cover a limited scale range, resulting in poor performance if the objects' scales mismatches with the receptive field. To achieve the goal of detecting all objects with SiSo encoders, we have to find a way to generate an output feature with various receptive fields, compensating for the lack of multiple-level features.

We begin with enlarging the receptive field of the C5 feature by stacking standard and dilated convolutions~\cite{yu2015multi}. Although the covered scale range is enlarged to some extent, it still can not cover all object scales as the enlarging process multiplies a factor greater than $1$ to all originally covered scales. We illustrate the situation in Figure~\ref{fig4}(b), where the whole scale range shifts to larger scales compare with the one in Figure~\ref{fig4}(a). Then, we combine the original scale range and the enlarged scale range by adding the corresponding features, resulting in an output feature with multiple receptive fields covering all object scales (Figure~\ref{fig4}(c)). The above operations can be easily achieved by constructing residual blocks~\cite{he2016deep} with dilations on the middle $3\times3$ convolution layer.

\begin{figure}
\centering
\includegraphics[width=.475\textwidth]{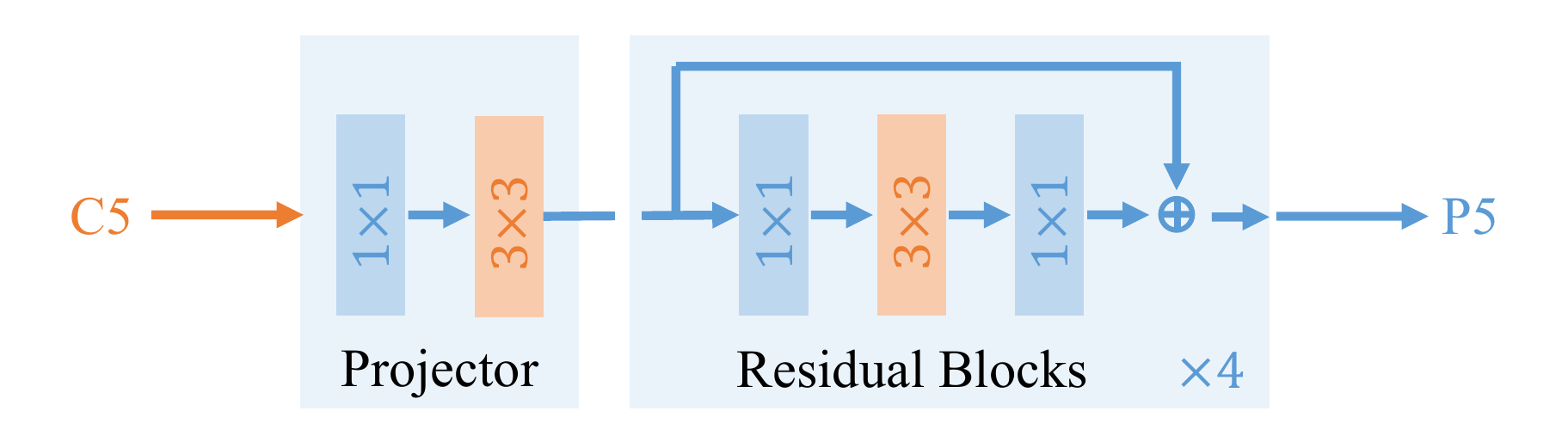}
\caption{An illustration of the structure of {\em Dilated Encoder}. In the figure, $1\times1$ and $3\times3$ denotes $1\times1$ and $3\times3$ convolution layers and $\times4$ means four successive residual blocks. All convolution layers in Residual Blocks are followed by a batchnorm layer~\cite{ioffe2015batch} and a ReLU layer~\cite{nair2010rectified}, while in Projector, we only use convolution layers and batchnorm layers~\cite{ioffe2015batch}.}
\label{fig5}\vspace{-2mm}
\end{figure}

\paragraph{Dilated Encoder:} Based on the above designs, we propose our SiSo encoder in Figure~\ref{fig5}, named as {\em Dilated Encoder}. It contains two main components: the {\em Projector} and the {\em Residual Blocks}. The projection layer first applies one $1\times1$ convolution layer to reduce the channel dimension, then add one $3\times3$ convolution layer to refine semantic contexts, which is the same as in the FPN~\cite{lin2017feature}. After that, we stack four successive dilated residual blocks with different dilation rates in the $3\times3$ convolution layers to generate output features with multiple receptive fields, covering all objects' scales. 

\paragraph{Discussion:} Dilated convolution~\cite{yu2015multi} is a common strategy to enlarge the features' receptive field in object detection. As reviewed in the Section~\ref{sec2}, TridentNet~\cite{li2019scale} use dilated convolution to generate multi-scale features. It deals with the scale variation problem in object detection via multi-branch structure and weight sharing mechanism, which is different from our single-level feature setting. Moreover, {\em Dilated Encoder} stack dilated residual blocks one by one without weight sharing. Although DetNet~\cite{li2018detnet} also successively applies dilated residual blocks, its purpose is to maintain the spatial resolution of the features and keep more details in the backbone's outputs, while ours is to generate a feature with multiple receptive fields out of the backbone. The design of {\em Dilated Encoder} enables us to detecting all objects on single-level feature instead of on multiple-level features like TridentNet~\cite{li2019scale} and DetNet~\cite{li2018detnet}.

\subsection{Imbalance Problem on Positive Anchors} \label{sec4.2}
The definition of positive anchors is crucial for the optimization problem in object detection. In anchor-based detectors, strategies to define positive are dominated by measuring the IoUs between anchors and ground-truth boxes. In RetinaNet~\cite{lin2017focal}, if the max IoU of the anchor and ground-truth boxes is greater than a threshold $0.5$, this anchor will be set as positive. We call it Max-IoU matching. 

\begin{figure}
\centering
\includegraphics[width=.475\textwidth]{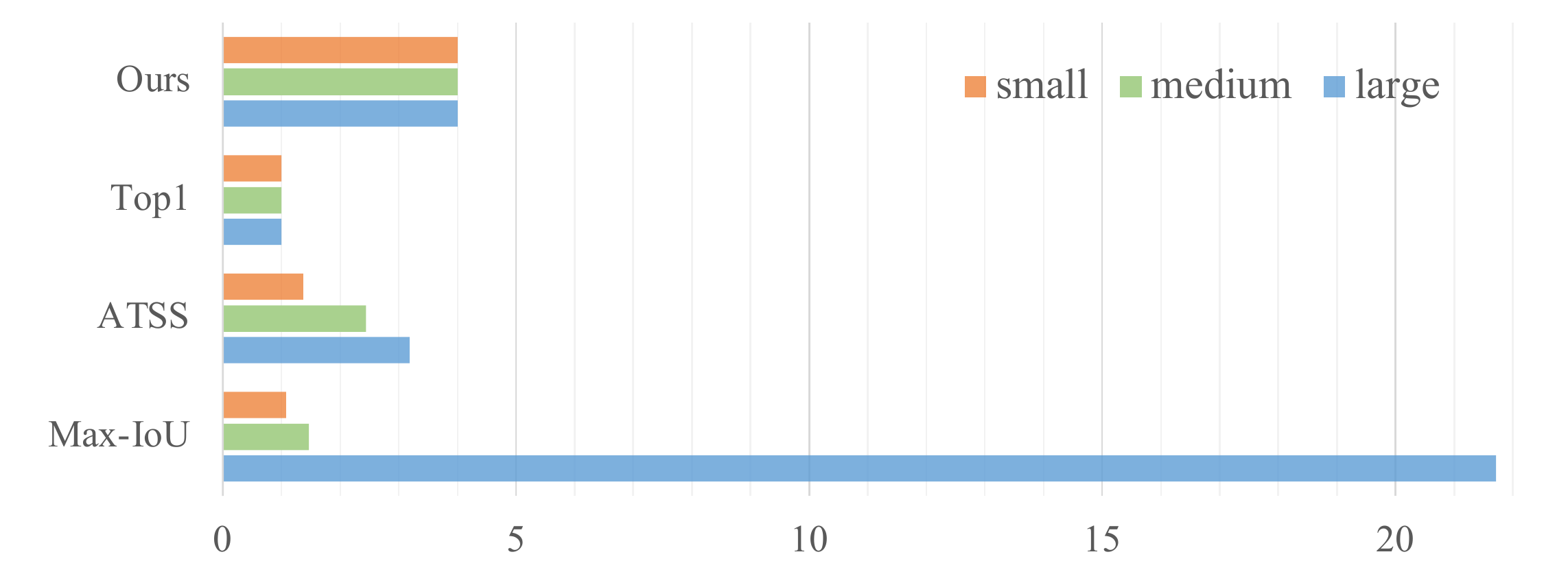}
\caption{Distribution of the generated positive anchors in various matching methods with single feature. This figure aims to show the balancedness of the generated positive anchors. The positive anchors in the Max-IoU are dominated by large ground-truth boxes, causing huge imbalance across object scales. ATSS alleviates the imbalance problem by adaptively sampling positive anchors when training. The Top1 and Ours adopt a uniform matching, generating positive anchors in a balanced manner regardless of small, medium, and large objects.}
\label{fig7}\vspace{-2mm}
\end{figure}

In MiMo encoders, the anchors are pre-defined on multiple levels in a dense paved fashion, and the ground-truth boxes generate positive anchors in feature levels corresponding to their scales. Given the divide-and-conquer mechanism, Max-IoU matching enables ground-truth boxes in each scale to generate a sufficient number of positive anchors. However, when we adopt the SiSo encoder, the number of anchors diminish extensively compare to the one in the MiMo encoder, from $100k$ to $5k$, resulting in sparse anchors\footnote{In SiSo encoders, we simply collapse multiple anchors on multiple-level features to single-level, e.g., we construct $5$ anchors with different anchor sizes of \{32, 64, 128, 256, 512\} on each position of the C5 feature.}. Sparse anchors raise a matching problem for detectors when applying Max-IoU matching, as shown in Figure~\ref{fig7}. Large ground-truth boxes induce more positive anchors than small ground-truth boxes in natural, which cause an imbalance problem for positive anchors. This imbalance makes detectors pay attention to large ground-truth boxes while ignoring the small ones when training. 

\paragraph{Uniform Matching:} To solve this imbalance problem in positive anchors, we propose an {\em Uniform Matching} strategy: adopting the {\em k nearest} anchor as positive anchors for each ground-truth box, which makes sure that all ground-truth boxes can be matched with the same number of positive anchors uniformly regardless of their sizes (Figure~\ref{fig7}). Balance in positive samples makes sure that all ground-truth boxes participate in training and contribute equally. Besides, following Max-IoU matching~\cite{lin2017focal}, we set IoU thresholds in Uniform Matching to ignore large IoU ($\textgreater0.7$) negative anchors and small IoU (\textless0.15) positive anchors.

\paragraph{Discussion: relation to other matching methods.} Applying topk in the matching process is not new. ATSS~\cite{zhang2020bridging} first select topk anchors for each ground-truth box on $\mathcal{L}$ feature levels, then samples positive anchors among $k\times\mathcal{L}$ candidates by dynamic IoU thresholds. However, ATSS focuses on defining positives and negatives adaptively, while our uniform matching focuses on achieving \textbf{balance on positive samples with sparse anchors}. Although several previous methods achieve balance on positive samples, their matching processes are \textbf{not} designed for this imbalance problem. For example, YOLO~\cite{redmon2016you} and YOLOv2~\cite{redmon2017yolo9000} match the ground-truth boxes with the best matching cell or anchor; DETR~\cite{carion2020end} and ~\cite{stewart2016end} apply Hungarian algorithm~\cite{kuhn1955hungarian} for matching. These matching methods can be view as top1 matching, which is a specific case of our uniform matching. More importantly, the difference between the uniform matching and the learning-to-match methods is that: the learning-to-match methods, such as FreeAnchor~\cite{zhang2019freeanchor} and PAA~\cite{paa-eccv2020}, adaptively separate anchors into positives and negatives according to the learning status, while uniform matching is \textbf{fixed} and does not evolve with training. The uniform matching is proposed to address the specific imbalance problem on positive anchors under the SiSo design. The comparison in Figure~\ref{fig7} and the results in Table~\ref{tab5e} demonstrate the significance of the balance in positives in SiSo encoders.

\subsection{YOLOF} \label{yolof}
Based on the solutions above, we propose a fast and straightforward framework with single-level feature, denoted as YOLOF. We format YOLOF into three parts: the backbone, the encoder, and the decoder. The sketch of YOLOF is shown in Figure~\ref{fig10}. In this section, we give a brief introduction to the main components of YOLOF. 

\paragraph{Backbone.} In all models, we simply adopt the ResNet~\cite{he2016deep} and ResNeXt~\cite{xie2017aggregated} series as our backbone. All models are pre-trained on ImageNet. The output of the backbone is the C5 feature map which has 2048 channels and with a downsample rate of 32. To make a fair comparison with other detectors, all batchnorm layers in the backbone are frozen by default.

\paragraph{Encoder.} For the encoder (Figure~\ref{fig5}), we first follow FPN by adding two projection layers (one $1\times1$ and one $3\times3$ convolution) after the backbone, resulting in a feature map with $512$ channels. Then, to enable the encoder's output feature to cover all objects on various scales, we propose to add residual blocks, which consist of three consecutive convolutions: the first $1\times1$ convolution apply channel reduction with a reduction rate of $4$, then a $3\times3$ convolution with dilation is used to enlarge the receptive field, at last, a $1\times1$ convolution to recover the number of channels. 
\begin{table*}[t]
\tablestyle{3.5pt}{1.47}
\begin{tabular}{cx{22}x{22}x{22}x{22}x{22}x{22}x{22}x{22}x{22}x{22}}
Model    &schedule    &$AP$   &$AP_{50}$     &$AP_{75}$    &$AP_{S}$   &$AP_{M}$   &$AP_{L}$ &\#params  &GFLOPs &FPS \\
\shline
RetinaNet~\cite{lin2017focal}  &1x     &35.9   &55.7     &38.5   &19.4   &39.5   &48.2 &38M           &201 &13 \\
RetinaNet-R101~\cite{lin2017focal}   &1x  &38.3   &58.5     &41.3   &21.7   &42.5   &51.2 &57M  &266  &11 \\
\hline
RetinaNet$+$        &1x   &37.7   &58.1     &40.2   &22.2   &41.7   &49.9 &38M      &201    &13 \\
RetinaNet-R101$+$  &1x  &40.0   &60.4     &42.7   &23.2   &44.1   &53.3 &57M    &266   &10\\
\hline
YOLOF             &1x   &37.7   &56.9     &40.6   &19.1   &42.5   &53.2  &44M     &86   &32\\
YOLOF-R101   &1x  &39.8   &59.4     &42.9   &20.5   &44.5   &54.9 &63M      &151   &21\\
YOLOF-X101   &1x   &42.2   &62.1     &45.7   &23.2   &47.0   &57.7   &102M   &289   &10\\
YOLOF-X101$^\dagger$    &3x   &44.7   &64.1     &48.6   &25.1   &49.2   &\textbf{60.9} &102M       &289   &10\\
YOLOF-X101$^{\dagger\ddagger}$ &3x &\textbf{47.1}   &\textbf{66.4}     &\textbf{51.2}   &\textbf{31.8}   &\textbf{50.9}   &60.6  &102M    &-   &-\\
\hline
\end{tabular}\vspace{2mm}
\caption{ Comparison with RetinaNet on the COCO2017 validation set. The top section shows the results of RetinaNet. The middle section gives the results of an improved RetinaNet (with a "$+$"), which is RetinaNet with GIoU~\cite{rezatofighi2019generalized}, GN~\cite{wu2018group}, and implicit objectness. The last section shows the results of various YOLOF models. In the table, the model with a suffix of R101 or X101 means it use ResNet-101~\cite{he2016deep} or RetNeXt-101-64$\times$4d~\cite{xie2017aggregated} as backbone. For those not marked with suffix, they adopt ResNet-50~\cite{he2016deep} by default. In the last two rows, we use multi-scale training and testing techniques ($\dagger$ indicates multi-scale training and $\ddagger$ means multi-scale testing), whose settings follow HTC~\cite{chen2019hybrid}. More details about the settings can be found in the Appendix. In the last three columns, we show models' number of parameters (\#params), GFLOPs, and inference speed. All FLOPs are measured with a shorter edge size 800 over the first 100 images of COCO val2017. Moreover, the FPS in the table is calculated with batch size 1 on 2080Ti from the total inference pure compute time reported in the Detectron2~\cite{wu2019detectron2}.}
\label{tab1}\vspace{-1mm}
\end{table*}
\paragraph{Decoder.} For the decoder, we adopt the main design of RetinaNet, which consists of two parallel task-specific heads: the classification head and the regression head (Figure~\ref{fig10}). We only add two minor modifications. The first one is that we follow the design of FFN in DETR~\cite{carion2020end} and make the number of convolution layers in two heads different. There are four convolutions followed by batch normalization layers and ReLU layers on the regression head while only have two on the classification head. The second is that we follow Autoassign~\cite{zhu2020autoassign} and add an implicit objectness prediction (without direct supervision) for each anchor on the regression head. The final classification scores for all predictions are generated by multiplying the classification output with the corresponding implicit objectness.

\paragraph{Other Details.} As mentioned in the previous section, the pre-defined anchors in YOLOF are sparse, decreasing the match quality between anchors and ground-truth boxes. We add a random shift operation on the image to circumvent this problem. The operation shifts the image randomly with a maximum of 32 pixels in left, right, top, and bottom directions and aims to inject noises into the object's position in the image, increasing the probability of ground-truth boxes matching with high-quality anchors. Moreover, we found that a restriction on the anchors' center's shift is also helpful to the final classification when using a single-level feature. We add a restriction that the centers' shift for all anchors should smaller than $32$ pixels.

\section{Experiments}
We evaluate our YOLOF on the MS COCO~\cite{lin2014microsoft} benchmark and conduct comparisons with RetinaNet~\cite{lin2017focal} and DETR~\cite{carion2020end}. Then, we provide a detailed ablation study of each component's design with quantitative results and analysis. Finally, to give insights to further research on single-level detection, we provide error analysis and show the weaknesses of YOLOF compared with DETR~\cite{carion2020end}. The details are as follows.

\paragraph{Implementation Details.} YOLOF is trained with synchronized SGD over 8 GPUs with a total of 64 images per mini-batch (8 images per GPU). All models are trained with an initial learning rate of $0.12$. Moreover, following DETR~\cite{carion2020end}, we set a smaller learning rate for the backbone, which is $1/3$ of the base learning rate. To stabilize the training at the beginning, we extend the number of warmup iterations from $500$ to $1500$. For training schedules, as we increase the batch size, the '$1\times$' schedule setting in YOLOF is a total of $22.5k$ iterations and with base learning rate decreased by 10 in the $15k$ and the $20k$ iteration. Other schedules are adjusted according to the principles in Detectron2~\cite{wu2019detectron2}. For model inference, we employ NMS with a threshold of $0.6$ to post-process the results. For other hyperparameters, we follow the settings of RetinaNet~\cite{lin2017focal}.
\begin{table*}[t]
\tablestyle{3.5pt}{1.47}
\begin{tabular}{cx{22}x{25}x{40}x{22}x{22}x{22}x{22}x{22}x{22}}
Model                   &Epochs    &\#params  &GFLOPS/FPS    &$AP$   &$AP_{50}$     &$AP_{75}$    &$AP_{S}$   &$AP_{M}$   &$AP_{L}$\\
\shline
DETR~\cite{carion2020end}              &500   &41M &86/24$^*$        &42.0   &62.4     &44.2   &20.5   &45.8   &61.1     \\
DETR-R101~\cite{carion2020end}    &500  &60M &152/17$^*$        &43.5   &\textbf{63.8}     &46.4   &21.9   &48.0   &\textbf{61.8}     \\
\hline
YOLOF             &72    &44M     &86/32     &41.6   &60.5     &45.0   &22.4   &46.2   &57.6     \\
YOLOF-R101   &72    &63M      &151/21     &\textbf{43.7}   &62.7     &\textbf{47.4}   &\textbf{24.3}   &\textbf{48.3}   &58.9     \\
\hline
\end{tabular}\vspace{2mm}
\caption{Comparison with DETR on the COCO2017 validation set. We conduct comparisons with backbone ResNet-50 (without suffix) and ResNet-101 (with a suffix R101). To make fair comparison, YOLOF adopts multi-scale training (same as in Table~\ref{tab1}) with a '6$\times$' schedule, which is roughly 72 epochs. For the FPS of DETR, $*$ means we follow the method in the original paper~\cite{carion2020end} and re-measure it on 2080Ti. }
\label{tab2}\vspace{-3mm}
\end{table*}

\subsection{Comparison with previous works}
\paragraph{Comparison with RetinaNet:} To make a fair comparison, we align RetinaNet with YOLOF by employing generalized IoU~\cite{rezatofighi2019generalized} for the box loss, adding an implicit objectness prediction, and applying group normalization layers~\cite{wu2018group} in heads (as there are only two images per GPU and both BN~\cite{ioffe2015batch} and SyncBN~\cite{zhang2018context} give poor results in RetinaNet~\footnote{\url{https://github.com/facebookresearch/detectron2/blob/master/detectron2/modeling/meta_arch/retinanet.py\#L532}}, we use GN~\cite{wu2018group} instead of BN~\cite{ioffe2015batch} in the heads). The results are presented in Table~\ref{tab1}. All '$1\times$' models are trained with a single scale that the shorter side is set as 800 pixels and the longer side is at most 1333~\cite{lin2017focal}. In the top section, we give RetinaNet baseline results trained with Detectron2~\cite{wu2019detectron2}. In the middle section, we present the results of the improved RetinaNet baseline (with a "$+$"), whose settings are aligned with YOLOF. In the last section, we show results from multiple YOLOF models. Thanks to the single-level feature, YOLOF achieves results {\em on par with} RetinaNet$+$ with a $57\%$ flops reduction (flops for each component in YOLOF are shown in Figure~\ref{fig3}) and a $2.5\times$ speed up. Due to the large stride (32) of the C5 feature, YOLOF has an inferior performance ($-3.1$) than RetinaNet$+$ on small objects. However, YOLOF achieves better performance on large objects (+3.3) as we add dilated residual blocks in the encoder. The comparison between RetinaNet$+$ and YOLOF with a ResNet-101~\cite{he2016deep} show similar evidence as well. Although YOLOF is inferior to RetinaNet$+$ on small objects when applying the same backbone, it can match small objects' performance with a stronger backbone ResNeXt~\cite{xie2017aggregated} while running at the same speed. Moreover, to prove that our method is compatible and complementary to current technologies in object detection, we show results that training with multi-scale images and a longer schedule in the last two rows of Table~\ref{tab1}. Finally, with the help of multi-scale testing, we obtain our final result of $47.1$ mAP and a competitive performance of $31.8$ mAP on small objects.
\begin{table}[t]
\tablestyle{1.8pt}{1.2}
\begin{tabular}{x{45}x{20}x{20}x{20}x{20}x{20}x{20}x{20}x{20}}
Model        &Epochs    &FPS    &$AP$  &$AP_{50}$    &$AP_{75}$    &$AP_{S}$   &$AP_{M}$   &$AP_{L}$\\
\shline
YOLOv4~\cite{bochkovskiy2020yolov4}    &273 &53$^*$        &43.5   &\textbf{65.7}   &47.3   &\textbf{26.7}   &47.6   &53.3     \\
YOLOF-DC5      &184 &\textbf{60$^\dagger$}        &\textbf{44.3}	 &62.9	&\textbf{47.5}   &24.0   &\textbf{48.5}   &\textbf{60.4}     \\
\hline
\end{tabular}\vspace{2mm}
\caption{ Comparison with YOLOv4 on the COCO {\em test-dev} set. We train YOLOF-DC5 with a '$15\times$' schedule (184 epochs) and compare it with YOLOv4. In the table, $\dagger$ means that the FPS for YOLOF-DC5 is measured by following YOLOv4~\cite{bochkovskiy2020yolov4}. It is different from the method used in Table~\ref{tab1}, \ref{tab2} in this paper. In YOLOv4~\cite{bochkovskiy2020yolov4}, the authors fuse the convolution layer and the batch normalization layer, then measure the inference time after converting the model to half-precision. $*$ represents that we get the speed for YOLOv4 on 2080Ti from the official repo \url{https://github.com/AlexeyAB/darknet\#geforce-rtx-2080-ti}.}
\label{tab3}\vspace{-2mm}
\end{table}
\paragraph{Comparison with DETR.}
DETR~\cite{carion2020end} is a recent proposed detector which introduces transformer~\cite{vaswani2017attention} to object detection. It achieves surprising results on the COCO benchmark~\cite{lin2014microsoft} and proves that by only adopting a single C5 feature, it can achieve comparable results with a multi-level feature detector (Faster R-CNN w/ FPN~\cite{lin2017feature}) for the first time. Given this, one might expect that layers capture global dependencies such as transformer layers~\cite{vaswani2017attention} are required to achieve promising results in single-level feature detection. {\em However, we show that a conventional network with local convolution layers can also achieve this goal}. We compare DETR with global layers and YOLOF with local convolution layers in Table~\ref{tab2}. The results show that YOLOF matches the DETR's performance, and YOLOF gets more benefits from deeper networks than DETR (w/ ResNet-50 ($-0.4$) {\em vs.} w/ ResNet-101 (+0.2)). Interestingly, we find that YOLOF outperforms DETR on small objects (+1.9 and +2.4) while lags behind DETR on large objects (-3.5 and -2.9). The finding is consistent with the local and global discussion above. More importantly, compared with DETR, YOLOF converge much faster ($\sim7\times$), making it more suitable than DETR to serve as a simple baseline for single-level detectors.
\begin{table}[t]
\tablestyle{1.8pt}{1.2}
\begin{tabular}{x{30}x{35}|x{22}x{22}|x{22}x{22}x{22}}
\tabincell{c}{{\em Dilated}\\{\em Encoder}}    &\tabincell{c}{{\em Uniform}\\{\em Matching}}  &$AP$     &$\Delta $    &$AP_{S}$   &$AP_{M}$   &$AP_{L}$\\
\shline
&           &21.1     &-16.6   &8.6    &31.1   &34.5     \\
\checkmark     &           &29.1    &-8.6   &9.5    &32.2   &50.6     \\
&\checkmark &33.8    &-3.9   &17.7   &40.9   &43.8     \\
 \checkmark     &\checkmark &\textbf{37.7}    &-   &\textbf{19.1}   &\textbf{42.5}   &\textbf{53.2}     \\
\hline
\end{tabular}\vspace{2mm}
\caption{ Effect of {\em Dilated Encoder} and {\em Uniform Matching} with ResNet-50. These two components improve the original single-level detector by 16.6 mAP. Note that the result of 21.1 mAP in the table is not a bug. It perform slightly worse than the detectors with SiSo encoders in Figure~\ref{fig1} and Figure~\ref{fig3} due to the design of the decoder in YOLOF - only two convolution layers in the classification head.}
\label{tab4}\vspace{-5mm}
\end{table}

\paragraph{Comparison with YOLOv4.} YOLOv4~\cite{bochkovskiy2020yolov4} is an optimal speed and accuracy multi-level feature detector. It combines many tricks to achieve state-of-the-art results. As our purpose is to build a simple and fast baseline for single-level detectors, investigation on the bag of freebie tricks is outside of the scope of this work. Thus, we do not expect a rigidly aligned comparison on performance. To compare our YOLOF with YOLOv4, we apply the data augmentation methods as YOLOv4, adopt a three-phase training pipeline, modify the training settings accordingly, and add dilations on the last stage of the backbone (YOLOF-DC5 in Table~\ref{tab3}). More technical details about the model and the training settings are given in the Appendix. As shown in Table~\ref{tab3}, YOLOF-DC5 can run $13\%$ faster than YOLOv4 with a $0.8$ mAP improvement on overall performance. YOLOF-DC5 achieves less competitive results on small objects than YOLOv4 ($24.0$ mAP vs. $26.7$ mAP) while outperforms it on large objects by a large margin ($+7.1$ mAP). The above results indicate that single-level detectors have great potential to achieve state-of-the-art speed and accuracy simultaneously.
\begin{table*}[t]\vspace{-3mm}
\subfloat[\textbf{Number of ResBlocks} (ResNet-50): More residual blocks bring more gains. {\em N} represent the number of ResBlocks. To keep YOLOF simple and neat, we add 4 blocks in the encoder by default.\label{tab5a}]{
\tablestyle{3pt}{1.05}\begin{tabular}{c|x{22}x{22}x{22}x{22}}
 \scriptsize {\em N} & AP & AP$_{s}$ & AP$_{m}$ & AP$_{l}$\\
\shline
 \scriptsize 0 & 33.8 & 17.7 & 40.9 & 43.8\\
 \scriptsize 2 & 34.9 &17.8	  & 41.3 &46.8\\
 \scriptsize \textbf{4} & \textbf{35.5} & \textbf{17.6}  & \textbf{41.4} & \textbf{48.4}\\
 \scriptsize 6 & 36.0 &17.7	 &41.9 &49.5\\
 \scriptsize 8 & 36.6 &18.5	 &42.0 &50.7\\
 \scriptsize 10 & 36.9 &18.3	 &42.4 &50.4
\end{tabular}}\hspace{3mm}
\subfloat[\textbf{Different dilations} (ResNet-50-N4): 'N4' means we add 4 ResBlocks in the encoder. Dilation in the residual block gives large gains on large objects and slightly improve the performance of small and medium objects.\label{tab5b}]{
\tablestyle{4pt}{1.05}\begin{tabular}{c|x{22}x{22}x{22}x{22}}
 \scriptsize {\em Dilations} & AP & AP$_{s}$ & AP$_{m}$ & AP$_{l}$\\
\shline
 \scriptsize 1,1,1,1 & 35.5 & 17.6 & 41.4 & 48.4\\
 \scriptsize 2,2,2,2 & 36.4 & 18.1 & 41.8 & 50.2\\
 \scriptsize 3,3,3,3 & 36.9 & 18.4 & 42.1 & 51.0\\
 \scriptsize 1,2,3,4 & 37.4 & 18.6 & 42.6 & 51.8\\
 \scriptsize \textbf{2,4,6,8} & \textbf{37.7} & \textbf{19.1} & \textbf{42.5} & \textbf{53.2}\\
 \scriptsize 3,6,9,12 & 37.3 & 18.7 & 42.1 & 52.6
\end{tabular}}\hspace{3mm}
\subfloat[\textbf{Add shortcut or not} (ResNet-50): YOLOF results with shortcuts or not on various dilation settings. Shortcut brings considerable gains on all object scales and becomes more important when the dilations are adopted (+3.6 AP with dilations 2,4,6,8 vs. +2.9 AP when dilations are all ones).\label{tab5c}]{
\tablestyle{4pt}{0.892}\begin{tabular}{c|x{22}x{22}x{22}x{22}}
 \scriptsize {\em Dilations \& Shortcut} & AP & AP$_{s}$ & AP$_{m}$ & AP$_{l}$\\
\shline
 \scriptsize \tabincell{c}{2,4,6,8\\\checkmark} & \textbf{37.7} & \textbf{19.1} & \textbf{42.5} & \textbf{53.2}\\
 \hline
 \scriptsize \tabincell{c}{2,4,6,8\\-} & 34.1 &16.2	& 38.4 & 47.5\\
 \hline
 \scriptsize \tabincell{c}{1,1,1,1\\\checkmark} & 35.5 & 17.6 & 41.4 & 48.4\\
 \hline
 \scriptsize \tabincell{c}{1,1,1,1\\-} & 32.6 & 15.0 & 38.4 & 44.2\\
\end{tabular}}\hspace{-1mm}
\subfloat[\textbf{Number of positives} (ResNet-50-N4): Number of positive anchors in {\em Uniform Matching}. Increase the positive anchor for each ground-truth box can improve the performance while it saturates when too many positive anchors. We choose the top4 anchors in YOLOF which achieves best results.\label{tab5d}]{
\tablestyle{4pt}{1.2}\begin{tabular}{c|x{22}x{22}x{22}x{22}x{22}x{22}}
 {\em topk} & AP & AP$_{50}$ & AP$_{75}$ & AP$_{s}$ & AP$_{m}$ & AP$_{l}$\\[.1em]
\shline
 top1 &35.9	&55.6	&38.4	&17.5	&40.3	&50.2     \\
 top2 &37.2	&56.7	&39.9	&18.9	&41.6	&52.0     \\
 top3 &37.5	&\textbf{57.1}	&40.2	&18.6	&41.9	&52.5     \\
 \textbf{top4} &\textbf{37.7} &56.9	&\textbf{40.6}	&\textbf{19.1}	&\textbf{42.5}	&\textbf{53.2}    \\
 top5 &37.5	&56.7	&40.3	&18.1	&42	&\textbf{53.2}
\end{tabular}}\hspace{5mm}
\subfloat[\textbf{Uniform matching {\em vs}. other matchings} (ResNet-50-N4): Comparison with other matching methods. Uniform Matching achieve balance in positive anchors and get the best results among other matching methods, which is consistent with the comparison in Figure~\ref{fig7}. Note that '*' represents that we get the best result for ATSS~\cite{zhang2020bridging} when setting topk as 15. More details can be found in the Appendix.\label{tab5e}]{
\tablestyle{4pt}{1.2}\begin{tabular}{c|x{22}x{22}x{22}x{22}x{22}x{22}}
 {\em Matching Methods} & AP & AP$_{50}$ & AP$_{75}$ & AP$_{s}$ & AP$_{m}$ & AP$_{l}$\\[.1em]
\shline
Max-IoU Matching~\cite{lin2017focal}        &29.1   &45.9   &29.6   &9.5    &32.2   &50.6    \\
ATSS(topk=9)~\cite{zhang2020bridging}               &34.6	&54.3	&37.1	&17.7	&40.6	&46.9     \\
ATSS(topk=15)~\cite{zhang2020bridging}$^*$       &36.5	&55.9	&38.6	&18.1	&41.4	&50.8     \\
Hungarian Matching~\cite{carion2020end}      &35.8	&55.5	&38.3	&18.2	&39.9	&50.2    \\
\textbf{Uniform Matching}          &\textbf{37.7}	&\textbf{56.9}	&\textbf{40.6}	&\textbf{19.1}	&\textbf{42.5}	&\textbf{53.2}     \\
\end{tabular}}\vspace{2mm}
\caption{\textbf{Ablations}. We show ablation experiments for {\em Dilation Encoder} and {\em Uniform Matching} on COCO2017 val set with ResNet-50.}
\label{tab5}\vspace{-1mm}
\end{table*}

\subsection{Ablation Experiments}
We run a number of ablations to analyze YOLOF. We first provide an overall analysis of the two proposed components. Then, we show the ablation experiments on detailed designs of each component. Results are shown in Table~\ref{tab4},\ref{tab5} and discussed in detail next.

\paragraph{Dilated Encoder and Uniform Matching:} Table~\ref{tab4} shows that both {\em Dilated Encoder} and {\em Uniform Matching} are necessary to YOLOF and bring considerable improvements. Specifically, Dilated Encoder has a significant impact on large objects (43.8 {\em vs.} 53.2) and slightly improves the results of small and medium objects. The results indicate that the {\em limited scale range} is a severe problem in the C5 feature (Section~\ref{sec4.1}). Our Dilated Encoder provides a simple but effective solution to this problem. On the other side, the performance of small and medium objects drops significantly ($\sim10 AP$) without uniform matching, while the large objects' performance is only lightly affected. The finding is consistent with the {\em imbalance problem on positive anchors} analyzed in Section~\ref{sec4.2}. The positive anchors are dominated by large objects, resulting in poor results on small and medium objects. Finally, when we remove both Dilated Encoder and Uniform Matching, a single-level feature detector's performance drops back to $\sim20$ mAP like the results in Figure~\ref{fig1} and Figure~\ref{fig3}.

\paragraph{Number of ResBlock:} YOLOF stacks residual blocks in the SiSo encoder. The results in Table~\ref{tab5a} shows that stacking more blocks gives extensive improvements on large objects, which is due to the increment of the feature scale range. Although we observe continuous improvements with more blocks, we choose to add four residual blocks to keep YOLOF simple and neat.

\paragraph{Different dilations:} Following the analysis in Section~\ref{sec4.1}, to enable the C5 feature to cover large scales, we replace the standard $3\times3$ convolution layer in the residual blocks with its dilated counterpart. We show the results with different dilations in the residual blocks in Table~\ref{tab5b}. Applying dilations to residual blocks bring improvements to YOLOF, while the improvements are saturated when using too large dilations. We conjecture that the reason for this phenomenon is that dilations of $2,4,6,8$ are enough to match object scales in all images.

\paragraph{Add shortcut or not:} Table~\ref{tab5c} shows that shortcuts play an essential role in Dilated Encoder. The performance of all objects will drop significantly if we remove the shortcuts in residual blocks. According to Section~\ref{sec4.1}, shortcuts combine different scale ranges. A largely and densely paved scale range covered by the feature is the critical factor for detecting all objects in a single-level feature manner.

\paragraph{Number of positives:} A comparison among the number of induced positive anchors by ground-truth boxes is conducted in Table~\ref{tab5d}. Intuitively, more positive anchors can achieve better performance as the learning will be easier when given more samples. Thus, in our uniform matching manner, we empirically increase the number of positive anchors induced by each ground-truth box. As shown in Table~\ref{tab5d}, the hyper-parameter $k$ is very robust for the performance when $k$ is larger than 1, which may suggest that the most important is the uniform matching manner in YOLOF. We set $top4$ for our uniform matching as it is the best choice according to the results. 

\paragraph{Uniform matching {\em vs}. other matchings:} We compare the uniform matching with other matching strategies for YOLOF and show results in Table~\ref{tab5e}. The proposed uniform matching strategy can achieve the best results, compatible with the imbalance analysis in Figure~\ref{fig7}. It worth noting that the Hungarian matching strategy can be roughly treated as Top1 matching (Table~\ref{tab5d}) so that they get similar performance.
The difference between them is that an anchor will only match one object in Hungarian matching while the Top1 matching does not have this constraint, and the experiments show that this is not important. The original ATSS find that top9 anchors are the best choice, while we find top15 anchors are much better in the single-level feature detector. By using top15 anchors, ATSS achieves a good result of 36.5 mAP while still lags behind our uniform matching by a 1.2 mAP gap.
\begin{figure}
\centering
\includegraphics[width=.48\textwidth]{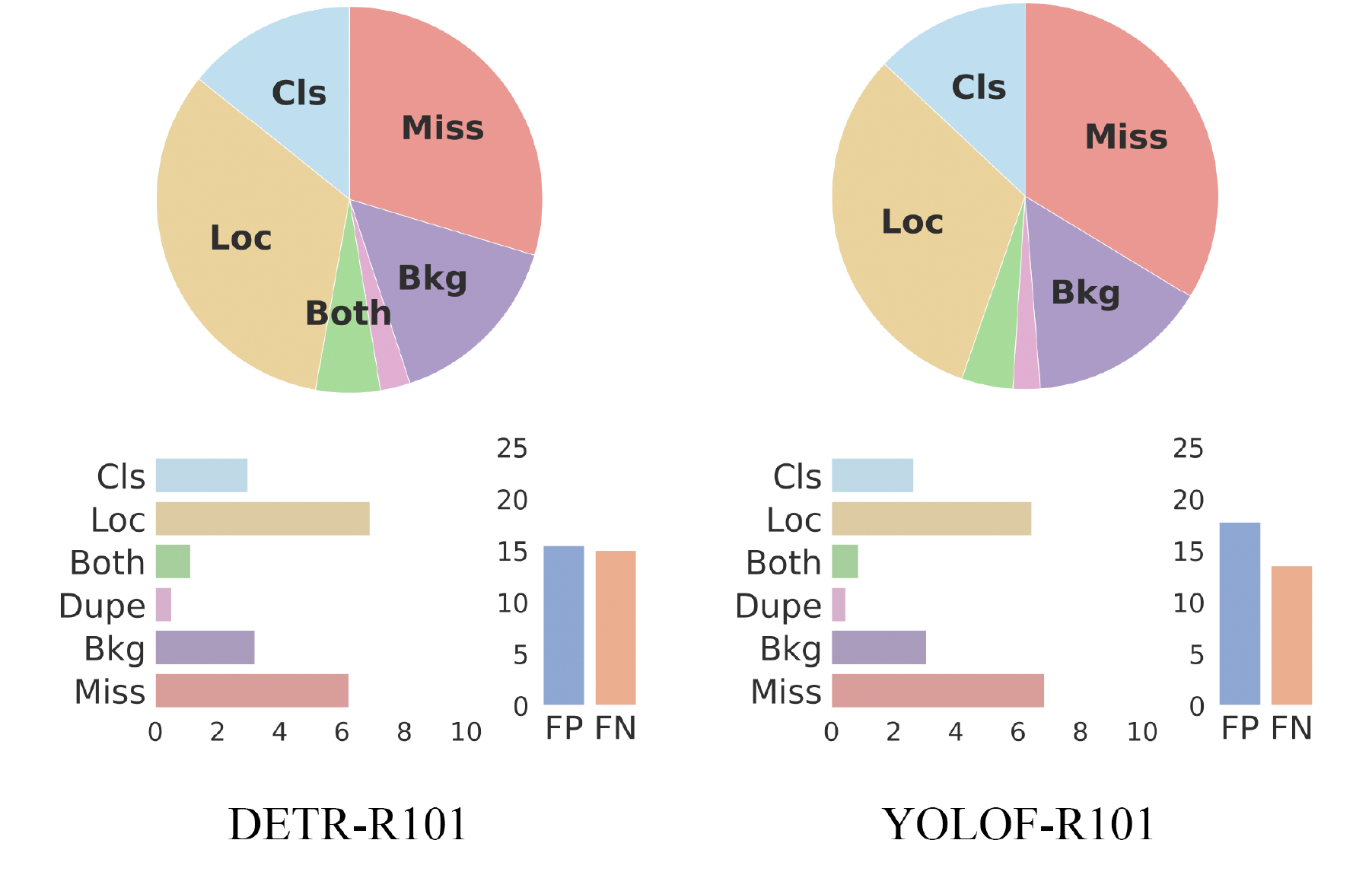}
\caption{Error analysis for DETR-R101 and YOLOF-R101. According to TIDE~\cite{bolya2020tide}, the figure shows the six types of errors (Cls: classification error; Loc: localization error; Both: both cls and loc error; Dupe: duplicate predictions error; Bkg: background error; Miss: missing error). The pie chart shows the relative contribution of each error, while the bar plots show their absolute contribution. FP and FN means false positive and false negative respectively.}
\label{fig8}\vspace{-3mm}
\end{figure}

\subsection{Error Analysis}
We add error analysis for YOLOF in this section to provide insights for future research in single-level feature detection. We adopt the recent proposed tool TIDE~\cite{bolya2020tide} to compare YOLOF with DETR~\cite{carion2020end}. As illustrated in Figure~\ref{fig8}, DETR has a larger error in localization than YOLOF, which may be related to its regression mechanism. DETR regresses objects in a total anchor free manner and predicts the location globally in the image, which causes difficulties in localization. In contrast, YOLOF relies on pre-defined anchors, which is responsible for higher missing error than DETR~\cite{carion2020end} in the predictions. According to the analysis in Section~\ref{sec4.2}, the anchors of YOLOF are sparsely and not flexible enough in the inference stage. Intuitively, there are situations that there are no high-quality anchors pre-defined around a ground-truth box. Thus, introducing the anchor-free mechanism into YOLOF may help alleviate this problem, and we leave it for future work. 
\begin{figure*}[ht]
\centering
\includegraphics[width=.95\textwidth]{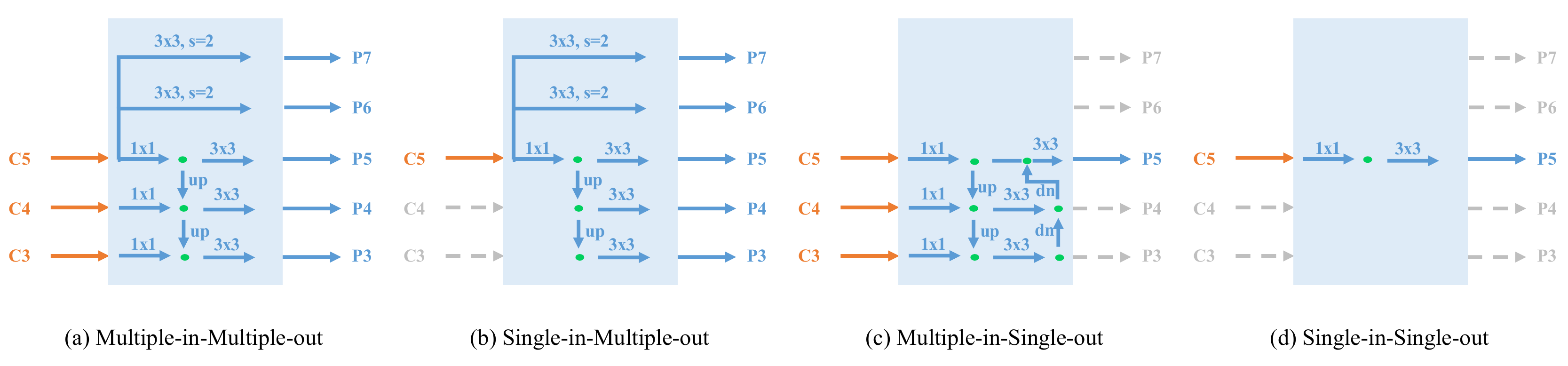}
\caption{Detailed Structures of Multiple-in-Multiple-out (MiMo), Single-in-Multiple-out (SiMo), Multiple-in-Single-out (MiSo), and Single-in-Single-out (SiSo) encoders.}
\label{fig9}
\end{figure*}
\begin{figure*}[ht]
\centering
\includegraphics[width=.95\textwidth]{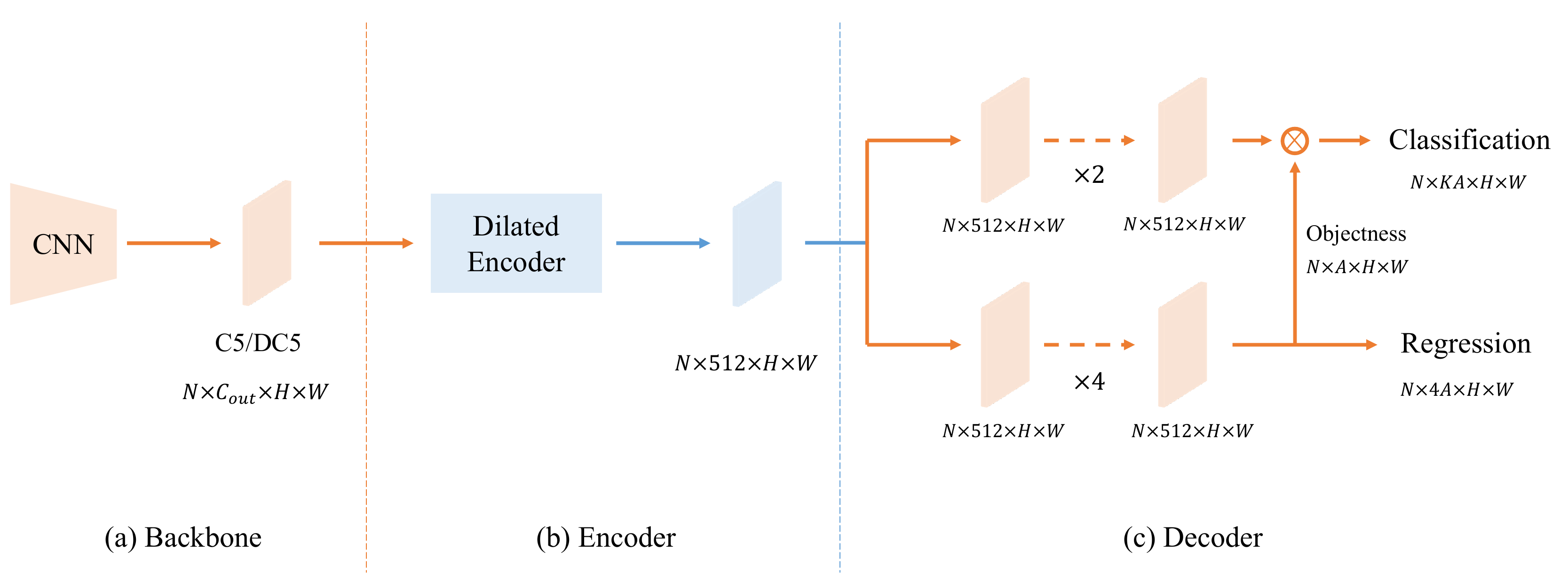}
\caption{The sketch of YOLOF, which consists of three main components: the backbone, the encoder, and the decoder. In the figure, 'C5/DC5' represents the output feature of the backbone with downsample rate of $32/16$. '$C_{out}$' means the number of channels of the feature. We set the number of channels as $512$ for feature maps in the encoder and the decoder. $H\times W$ is the height and width of feature maps.}
\label{fig10}
\end{figure*}

\section{Conclusion}
In this work, we identify that the success of FPN is due to its divide-and-conquer solution to the optimization problem in dense object detection. Given that FPN makes network structure complex, brings memory burdens, and slows down the detectors, we propose a simple but highly efficient method without using FPN to address the optimization problem differently, denoted as YOLOF. We prove its efficacy by making fair comparisons with RetinaNet and DETR. We hope our YOLOF can serve as a solid baseline and provide insight for designing single-level feature detectors in future research. 

\appendix
\section*{Appendix A: More Details} \label{Appendix:A}

\paragraph{Detailed Structures of All Encoders:}
In Figure~\ref{fig9}, we illustrate the detailed processes of generating outputs in encoders. The four encoders differ in the number of input features and output features. (a) The {\em Multiple-in-Multiple-out} (MiMo) encoder receives three levels from the backbone and output five levels. The structure of the MiMo encoder is the same as FPN in RetinaNet~\cite{lin2017focal}. (b) {\em Single-in-Multiple-out} (SiMo) only has one C5 feature for the input. As there are no other inputs, we remove the $1\times1$ convolution layer designed for C3 and C4. (c) {\em Multiple-in-Single-out} (MiSo) receives three input features while only generate one output feature P5. To fully utilize the context in the input features, we adopt a structure similar to PANet~\cite{liu2018path} in MiSo. (d) In the {\em Single-in-Single-out} (SiSo) encoder, we remove all other convolution layers and only keep the convolution layers in the level of C5.

\paragraph{Network Architecture of YOLOF}
In Figure~\ref{fig10}, we show a detailed network architecture of YOLOF. YOLOF detects objects on single-level feature, which is very simple. Our method consists of three components: the backbone, the encoder, and the decoder. The detailed design of these components are presented in Section~\ref{yolof}.

\paragraph{Training Time \& Memory:}
In this section, we compare training time and training memory among YOLOF, DETR~\cite{carion2020end}, and RetinaNet$+$~\cite{lin2017focal}. As shown in Table~\ref{tab7}, due to the long training schedule, DETR needs 112.5 hours to converge on COCO with eight 2080Ti GPUs, while YOLOF and RetinaNet$+$ only need 4.5 hours and 9.8 hours, respectively. As for training memory, YOLOF needs less memory than RetinaNet$+$ and DETR, which make YOLOF be trained with larger batch size and converge faster.
\begin{table}[t]
\tablestyle{1.8pt}{1.2}
\begin{tabular}{c|cc}
Model   &Memory/Images &Training Time\\ 
\shline
YOLOF &5.3G / \textbf{8} &4.5h\\
RetinaNet$+$~\cite{lin2017focal} &4.9G / 2 &9.8h\\
DETR~\cite{carion2020end} &7.1G / 2 &112.5h
\end{tabular}\vspace{2mm}
\caption{Comparison of training memory and training time among different models. All models are trained with eight 2080Ti GPUs with their default settings, i.e, we train YOLOF and RetinaNet$+$~\cite{lin2017focal} in a '1x' schedule, while train DETR~\cite{carion2020end} with 150 epochs on COCO2017 training set.}
\label{tab7}\vspace{-2mm}
\end{table}

\paragraph{More Implementation Details:}
The default training settings for YOLOF is a total of 64 images per mini-batch (8 images per GPU) with an initial learning rate of $0.12$. While for ResNeXt-101~\cite{xie2017aggregated}, we train with 4 images per GPU (batch size 32) and set the learning rate to $0.06$ following the linear rule~\cite{goyal2017accurate}. For multi-scale training, DETR~\cite{carion2020end} apply random crop plus resize to simulate large image size during training. In YOLOF, we simply resize the image to large size. For multi-scale training, we follow HTC~\cite{chen2019hybrid} and adopt a strategy of random sample the image size between $[400, 1400]$ with its largest edge no greater than 1600 pixels.

\paragraph{Detailed Settings to Compare with YOLOv4}
To match the performance of YOLOv4, we first increase the number of dilated residual blocks in the dilated encoder from $4$ to $8$. We adjust the dilations of these dilated residual blocks according to experimental results. We find that the dilations $[1, 2, 3, 4, 5, 6, 7, 8]$ give the best result. Then following YOLOv4~\cite{bochkovskiy2020yolov4}, we adopt its data augmentations, take the CSPDarkNet-53~\cite{wang2020cspnet} as the backbone, replace all the batch normalization layers with its synchronized counterpart, and apply LeakyReLU~\cite{xu2015empirical} in the encoder and the decoder instead of ReLU layers. According to the results in Table~\ref{tab9}, YOLOF-DC5 gives better results than YOLOF. Thus we use YOLOF-DC5 as the baseline model in this section. After that, we set an initial learning rate of $0.04$ for the whole model. To train the final model, we adopt a three-phase training. At first, we training YOLOF-DC5 for a '$9\times$' schedule; then we increase the ignore threshold for negative anchors from $0.75$ to $0.8$ and train a '$3\times$' schedule based on the previous model (this phase gives a $0.5$ mAP gain); at last, we train another '$3\times$' schedule by following the recipe introduced in~\cite{zhang2020swa}. The final result shown in Table~\ref{tab3} is produced by the SWA model, which is obtained by averaging $12$ checkpoints (the SWA model gives a $\sim 1$ mAP improvement).

\section*{Appendix B: Additional Experimental Results} \label{Appendix:B}

\paragraph{Number of Anchors:}
In RetinaNet~\cite{lin2017focal}, anchors are generated from multiple level features (P3-P7) with areas of $32^2$ to $512^2$, respectively. At each level feature, RetinaNet paves anchors with sizes $\{2^0, 2^{1/3}, 2^{2/3}\}$ and aspect ratios $\{0.5, 1, 2\}$. While in YOLOF, we only have a one-level feature to place anchors. To cover all objects' scales, we add anchors with areas of $\{32^2, 64^2, 128^2, 256^2, 512^2\}$, size $\{1\}$, and aspect ratio $\{1\}$ in the single feature map, resulting in 5 anchors in each position. Moreover, we investigate the influence of more anchors in YOLOF. Following RetinaNet, we generate 45 anchors in each position with different sizes ($\{2^0, 2^{1/3}, 2^{2/3}\}$) and more aspect ratios ($\{0.5, 1, 2\}$). All results are shown in Table~\ref{tab6}. The results show that adding more aspect ratios does not change the performance of YOLOF, while the performance drops with more sizes. Thus, we choose to add a minimum of five anchors for YOLOF by default.
\begin{table}[t]
\tablestyle{1.8pt}{1.2}
\begin{tabular}{x{30}|x{18}x{18}x{18}|x{18}x{18}x{20}|x{18}x{18}x{18}}
Model   &areas &sizes &ratios &$AP$   &$AP_{50}$  &$AP_{75}$  &$AP_{s}$   &$AP_{m}$   &$AP_{l}$\\ 
\shline
YOLOF &5 &1 &1 &\textbf{37.7} &56.9 &40.6 &19.1 &42.5 &53.2\\
YOLOF &5 &1 &3 &\textbf{37.7} &57.2 &40.7 &19.4 &42.0 &52.2\\
YOLOF &5 &3 &1 &35.0 &52.3 &37.9 &15.0 &40.6 &52.9\\
YOLOF &5 &3 &3 &35.4 &52.4 &38.3 &14.8 &41.2 &52.5
\end{tabular}\vspace{2mm}
\caption{Results of YOLOF with different multiple anchors per location on COCO~\cite{lin2014microsoft} validation set.}
\label{tab6}\vspace{-2mm}
\end{table}

\begin{table}[t]
\tablestyle{1.8pt}{1.2}
\begin{tabular}{c|x{15}|x{15}|x{15}|x{15}|x{15}|x{15}|x{15}|x{15}}
Model \& k   &5 &7 &9 &11 &13 &15 &17 &19\\ 
\hline
YOLOF (with ATSS) &33.7 &33.8 &34.6 &35.8 &35.5 &\textbf{36.5} &36.3 &36.2\\
\end{tabular}\vspace{2mm}
\caption{An illustration of how performance changes with the variation of the hyper-parameter $k$ in ATSS~\cite{zhang2020bridging}.}
\label{tab8}\vspace{-2mm}
\end{table}

\paragraph{Hyper-parameter of ATSS:}
Here, we provide the results of using different values of $k$ in ATSS~\cite{zhang2020bridging} in Table~\ref{tab8}. The results show that the choice of $k=9$ used in the original paper is not the best choice in YOLOF. According to the results, we choose $k=15$ for ATSS in this paper.

\begin{table}[t]
\tablestyle{1.8pt}{1.2}
\begin{tabular}{x{70}x{18}x{18}x{18}x{18}x{18}x{18}x{18}}
Model        &FPS    &$AP$  &$AP_{50}$    &$AP_{75}$    &$AP_{S}$   &$AP_{M}$   &$AP_{L}$\\
\shline
YOLOF-DC5-R50    &24      &39.2   &58.6   &42.7   &22.3   &43.9   &50.8     \\
YOLOF-DC5-R101$^*$      &17       &40.5	 &59.8	&43.9   &23.0  &44.9  &53.8    \\
\hline
\end{tabular}\vspace{2mm}
\caption{Additional results of YOLOF-DC5 with different backbones on COCO {\em val} split. $*$ means that due to the limited memory of 2080Ti, we train with 4 images per GPU (batch size 32) for ResNet-101. Higher performance can be achieved if train with 8 images per GPU or apply SyncBN (BN layers in the encoder and decoder restrict the improvements).}
\label{tab9}\vspace{-2mm}
\end{table}

\paragraph{Results with Dilated C5:}
In this paper, we show that YOLOF performs well on the C5 feature. To boost the performance of YOLOF, we detect objects on a feature map with higher resolution than the C5 feature. Following DETR~\cite{carion2020end}, we construct a backbone with dilation and without stride on its last stage. The backbone's output feature is denoted as DC5, with a downsample rate of 16. In Table~\ref{tab9}, we show the results of YOLOF-DC5 on COCO {\em val} split with ResNet-50 and ResNet-101 as the backbone. YOLOF-DC5 achieves higher performance than the original YOLOF but runs at a slower speed as the feature's resolution is larger than C5. To achieve the results, we first add a smaller anchor, resulting in $6$ anchors per location ($\{16, 32, 64, 128, 256, 512\}$), then we increase the $topk$ from $4$ to $8$ and change the ignore threshold for positive anchors from $0.15$ to $0.1$. Other parameters are the same as before.

\section*{Acknowledgements}
This work is supported by The National Key Research and Development Program of China (No. 2017YFA0700800), Beijing Academy of Artificial Intelligence (BAAI), National Natural Science Foundation of China (No.61972396, 61876182, 61906193), National Key Research and Development Program of China (No. 2020AAA0103402), the Strategic Priority Research Program of Chinese Academy of Sciences (No. XDB32050200), and The NSFC-General Technology Collaborative Fund for Basic Research (Grant No. U1936204).

{\small
\bibliographystyle{ieee_fullname}
\bibliography{cvpr}
}

\end{document}